\documentclass[runningheads]{llncs}
\usepackage{graphicx}
\usepackage{amsmath}
\usepackage{floatrow}
\newfloatcommand{capbtabbox}{table}[][\FBwidth]
\usepackage{blindtext}
\usepackage{multirow}

\begin{document}
\title{ConfNet2Seq: Full Length Answer Generation from Spoken Questions}
\titlerunning{ConfNet2Seq}

%\author{Anonymous author}
\author{Vaishali Pal\inst{1}\orcidID{0000-0002-1493-3659} \and
Manish Shrivastava\inst{1}\orcidID{0000-0001-8705-6637} \and Laurent Besacier\inst{2}\orcidID{0000-0001-7411-9125}}
\authorrunning{V. Pal et al.}
%\institute{Anonymous Institute}
\institute{LTRC - International Institute of Information Technology -  Hyderabad, India
\email{vaishali.pal@research.iiit.ac.in, m.shrivastava@iiit.ac.in} \\
\and LIG - Universit\'e Grenoble Alpes, France\\
\email{laurent.besacier@univ-grenoble-alpes.fr}}

\maketitle              % typeset the header of the contribution
%261,745
\begin{abstract}
Conversational and task-oriented dialogue systems aim to interact with the user using natural responses through multi-modal interfaces, such as text or speech. These desired responses are in the form of full-length natural answers generated over facts retrieved from a knowledge source. While the task of generating natural answers to questions from an answer span has been widely studied, there has been little research on natural sentence generation over spoken content. We propose a novel system to generate full length natural language answers from spoken questions and factoid answers. The spoken sequence is compactly represented as a confusion network extracted from a pre-trained Automatic Speech Recognizer. This is the first attempt towards generating full-length natural answers from a graph input(confusion network) to the best of our knowledge. We release a large-scale dataset of 259,788 samples of spoken questions, their factoid answers and corresponding full-length textual answers. Following our proposed approach, we achieve comparable performance with best ASR hypothesis.

%We also propose a graph-to-sequence based architecture to evaluate the efficacy of our dataset and approach.
%and show performance improvements by using confusion network over that of the best hypothesis of the ASR. 
%  
%We have used a hierarchical pointer-architecture across the spoken question represented as a confusion network and a conventional pointer-architecture across the textual factoid answer for generating the full-length answer. To achieve this task, we have generated a dataset of 300,000 samples of spoken SQuAD questions, factoid answers and full-length textual answers to train and evaluate our system. 
%We have achieved a BLEU score of and ROGUE score of on SQuAD dataset for this task.

\keywords{Confusion Network \and Pointer-Generator \and Copy Attention \and Natural Answer Generation \and Question Answering}
\end{abstract}

\section{Introduction}
\label{sec:intro}
Full-length answer generation is the task of generating natural answers over a question and an answer span, usually a fact-based phrase (factoid answer), extracted from relevant knowledge sources such as knowledge-bases (KB) or context passages. Such functionality is desired in conversational agents and dialogue systems to interact naturally with the user over multi-modal interfaces, such as speech and text. Typical task-oriented dialogue systems and chatbots formulate coherent responses from conversation context with a natural language generation (NLG) module. These modules copy relevant facts from context while generating new words, maintaining factual accuracy in a coherent fact-based natural response. Recent research \cite{Liu-et-al:0018ijcai,pal-etal-2019-answering} utilizes a pointer-network to copy words from relevant knowledge sources. While the task of generating natural response to text-based questions have been extensively studied, there is little research on natural answer generation from spoken content. Recent research on Spoken Question Answering and listening comprehension tasks\cite{DBLP:journals/corr/abs-1804-00320} extracts an answer-span and does not generate a natural answer. This motivates us to propose the task of generating full length answer from spoken question and textual factoid answer. However, such a task poses significant challenges as the performance of the system is highly dependent on Automatic Speech Recognizer (ASR) error. To mitigate the effect of Word Error Rate (WER) on ASR predictions, a list of top-N hypothesis, ASR lattices or confusion networks has been used in various tasks such as Dialogue-state-tracking \cite{DBLP:journals/corr/JagfeldV17,zhong2018global}, Dialogue-Act detection \cite{8462030} and named-entity recognition. These tasks show that models trained using multiple ASR hypotheses outperforms those trained top-1 ASR hypothesis. While classification and labeling tasks benefit from multiple hypothesis by aggregating the predictions over a list of ASR hypothesis, it is non-trivial to apply the same for NLG using pointer-networks. Our proposed system aims to take advantage of multiple time-aligned ASR hypotheses represented as a confusion network using a pointer-network to generate full-length answers. To the best of our knowledge, there is no prior work for full length answer generation from spoken questions. Our overall research contributions are as follows:
\begin{itemize}
  \item We propose a novel task of full-length answer generation from spoken question. To achieve this, we develop a ConfNet2Seq model which encodes a confusion network and adapts it over a pointer-generator architecture.
  \item We compare the effects of using multiple hypothesis encoded with a confusion network encoder and the best hypothesis encoded with a text encoder.
  \item We publicly release the dataset, comprising of spoken question audio file, the corresponding confusion network, the factoid answer and full-length answer.
\end{itemize}

\section{Related Work}
%Many spoken language classification systems use a list of top-K hypothesis instead of the best hypothesis. An alternative is to use a richer hypotheses space provided by the ASR lattice or a confusion network.
Spoken Language Understanding(SLU) has the additional challenge of disambiguation of ASR errors which drastically affect performance. Several methods have been proposed to curb the effects of the WER. Word lattices from ASR were first used by \cite{DBLP:journals/csl/Hakkani-TurBRT06} over ASR top-1 hypothesis for tasks such as named-entity extraction and call classification. Word confusion networks have been recently used by \cite{Ladhak2016LatticeRnnRN} for intent classification in dialogue systems and by \cite{DBLP:journals/corr/JagfeldV17,pal2020modeling} for dialogue state tracking (DST). \cite{DBLP:journals/corr/JagfeldV17} show that confusion network gives comparable performance to top-N hypotheses of ASR while \cite{pal2020modeling} show that using confusion network improves performance in both in time and accuracy. Another related task in SLU is that of Spoken Question Answering. Recent work \cite{DBLP:journals/corr/abs-1804-00320} on SQuAD dataset introduces the task for machine listening comprehension where the context passages are in audio form. \cite{DBLP:journals/corr/abs-1808-02280} released Open-Domain Spoken Question Answering Dataset (ODSQA) with more than three thousand questions in Chinese and used an enhanced word embedding comprising of word embedding and  pingyin-token embedding. \cite{unlu-inproceedings} developed a QA system for spoken lectures and generates an answer span from the video transcription. 

\section{Models}
\begin{figure}[h!]
 \centering
   \includegraphics[scale=0.36]{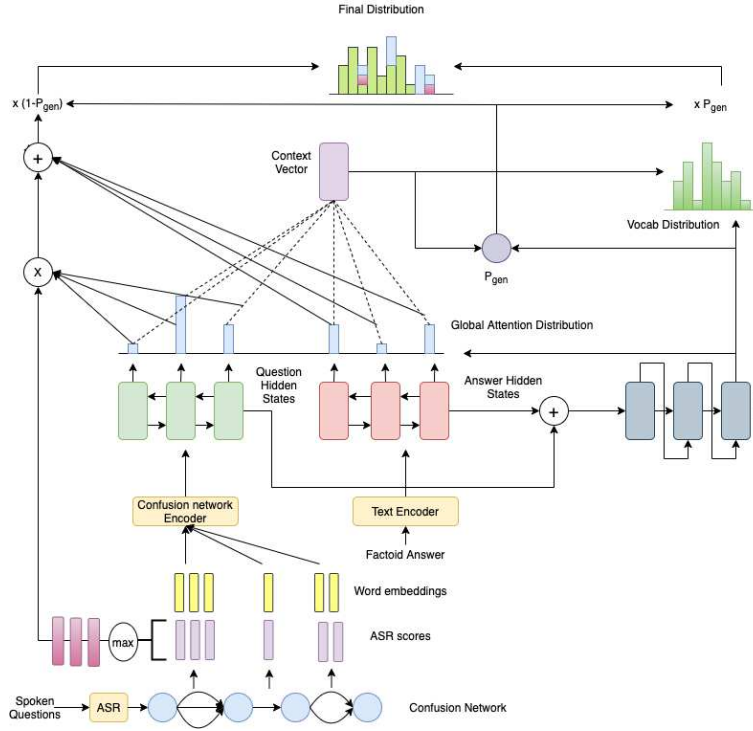}
 \caption{Full Length Answer Generation from Spoken Question and textual factoid answer: The confusion-network encoder generates a sequence of 1D-encodings of the sequence of parallel arcs(2D graph). The ASR scores are multiplied with the global-attention weights of the encodings to generate the copy-attention distribution of the question.}
 \label{fig:confnet_copy}
\end{figure}

Our system generates full length answer from a textual factoid answer and spoken question. We use a pointer generator architecture over two sequences, i.e., over the textual factoid answer sequence and the encoded question sequence produced by the confusion network encoder. In this section, we describe the 1) Confusion network encoder, 2) Final model over spoken question and factoid answer. The full architecture is shown in figure \ref{fig:confnet_copy}.

\subsection{Confusion Network Encoder}
\label{sec:conf_encoder}
A Confusion Network is a weighted directed acyclic graph with one or more parallel arcs between consecutive nodes where each path goes through all the nodes. Each set of parallel arcs represents time-aligned alternative words or hypothesis of the ASR weighed by probability. The total probability of all parallel arcs between two consecutive nodes sums up to 1. A confusion network $C$ can be defined formally as a sequence of sets of parallel weighted arcs as:
\begin{equation}
\begin{split}
    C=[(<w^1_1,\pi^1_1>,<w^2_1,\pi^2_1>,...,<w^{n_1}_1,\pi^{n_1}_1>),\ldots, \\ (<w^1_m,\pi^1_m>,<w^2_m,\pi^2_m>,...,
    <w^{n_m}_m,\pi^{n_m}_m>)]
\end{split}
\end{equation}
where $w^j_i$ is the $j^{th}$ ASR hypothesis at position $i$, and $\pi^j_i$ its associated probability. We use a confusion network encoder to transform a 2-dimensional confusion network into an 1-dimensional sequence of embeddings as described in \cite{masamura-et-al-2018}. Each word   $w^j_i$ of the confusion network can be encoded by weighing the word embedding by the ASR probability followed by a non-linear transformation as:
\begin{equation}
    q^j_i = \tanh(W_1 \pi^j_i Embedding(w^j_i))
\end{equation}
where $W_1$ is a trainable parameter. Each set of parallel arcs can be encoded into a vector by a weighted sum over the words of the parallel arc set. The weights measure the relevance of each word among the alternate time-aligned hypothesis. The learnt weight distribution for each parallel-arc set is:
\begin{equation}
     \alpha_i^j = \frac{\exp(W_2 q_i^j)} {\sum_j \exp(W_2 q_i^j)}
\end{equation}
where $W_2$ is a trainable parameter. The final encoding of each set of parallel arcs is: 
\begin{equation}
\label{eq:beta}
    \beta_i = \sum_i \alpha_i^j q_i^j
\end{equation}

\subsection{Full Length Answer Generation from Spoken Questions}
\label{sec:full_len}

We have followed a Seq2Seq with pointer generator architecture as \cite{pal-etal-2019-answering} to generate full-length answers from a question and factoid answer. However, we query with spoken questions instead of textual questions. The confusion network is extracted from spoken questions using a standard ASR. The question is encoded as $Q = \{q_{1}, q_{2}, ..., q_{n}\}$ where $q_t$ is the encoding from the confusion network encoder explained in section \ref{sec:conf_encoder}.

The factoid answer is represented as $A = \{ a_{1}, a_{2}, a_{3}, ..., a_{m} \}$ where $a_t$ is the GloVe embedding \cite{Pennington14glove:global} of a word. We encode the sequences using two 3-layered bi-LSTMs which share weights as:
\begin{equation}
\begin{split}
    h_{Q}^t = BILSTM(h_{Q}^{t-1}, q_{t}) \\
    h_{A}^t = BILSTM(h_{A}^{t-1}, a_{t})    
\end{split}
\end{equation}
The encoded hidden states of the 2 encoders are stacked together to produce a single list of source hidden states, $h_S =[h_{Q}; h_{A}]$. The decoder is initialized with the combined final states of the two encoders as $h_{T}^0 = h_{Q}^n + h_{A}^m$. 

The global attention weights  $attn_i^t$  are computed on the $n$ hidden states of the question and $m$ hidden states of the answer, stacked to produce a total of $m+n$ global attention weights. For each source state, $h_i$, and decoder state, $s_t$:
\begin{equation}
    \label{eq:global_attn}
    attn_i^t=softmax(v^{T}tanh(W_hh_i + W_ss_t+b_{attn}))  
\end{equation}
where $b_{attn}$, $v$, $W_h$, $W_s$ are learnable parameters. The copy mechanism for summarization introduced in \cite{DBLP:journals/corr/SeeLM17} takes advantage of a word distribution over an extended vocabulary comprising of source words and vocabulary words. The probability of copying a word $w$ from a text sequence is $\sum_{k:w_k=w}^m attn_k^t$.
To copy words from the confusion network, we compute the global attention weights over each set of  parallel-arc encodings. Here, the global attention weights denote a probability distribution over parallel-arc sets instead of words. These attention weights $attn_i^t$ are sampled to select the hidden state representation, $\beta_i$, of a set of parallel arcs. The ASR scores $\pi_i$ is a probability distribution over the set of parallel words at position $j$ in the confusion network. These are sampled to select the most likely word from that set of parallel arcs. The final probability of copying a word from the confusion network is the joint-probability:
\begin{equation}
    \widetilde{P}_{copy}(w) = \sum_{i:w_i^j=w}^n attn_i^t \pi_i^j
\end{equation}
The probability of copying a word from the answer is:
\begin{equation}
    P_{copy}(w) = \sum_{k:w_k=w}^m attn_k^t
\end{equation}
The final probability of a word output at $P(w)$ at time $t$ by the decoder is as shown in
\begin{equation}
     P(w) =P_{gen}P_{vocab}(w)+(1-P_{gen})(\widetilde{P}_{copy}(w) + P_{copy}(w)) 
\end{equation}
where $P_{gen}$ is a soft switch for the decoder to generate words or copy words from the source. $P_{vocab}(w)$ is the probability of generating a word from the vocabulary. These parameters are computed as described in \cite{DBLP:journals/corr/SeeLM17}.

\section{Dataset}
\label{lab:dataset}
%SQuAD is a machine reading comprehension dataset consisting of questions on Wikipedia articles, where the answer is a span from the corresponding passage related to the question.  We have used $300,000$ samples of answerable questions from the full-length answer generation dataset introduced in \cite{pal-etal-2019-answering}.
To generate data for our task, we use $258,478$ samples from the full-length answer generation dataset introduced in \cite{pal-etal-2019-answering} where each sample consists of a question, factoid answer and full-length answer. The samples in the dataset were chosen from SQuAD and HarvestingQA. Each sample in our dataset is also a 3-tuple $(q, f, a)$ in which $q$ is a spoken-form question, $f$ is a text-form factoid answer and $a$ is the text-form full-length natural answer. $256,478$ samples were randomly selected as the training set, $1000$ as the development set and $1000$ as the test set. We also extracted $470$ samples from NewsQA dataset and $840$ samples from Freebase to evaluate our system on cross-domain datasets.\footnote{Code and dataset at: https://github.com/kolk/ConfnetPointerGenBaseline}

We used Google text-to-speech to generate the spoken utterances of the questions. Google Voice \textit{en-US-Standard-B} was used to generate $239,746$ spoken questions in male voice and Google Voice
\textit{en-US-Wavenet-C} was used to generate $16,730$ spoken questions in female voice. All samples are in US accented English. The ASR lattice was extracted using Kaldi ASR \cite{Povey_ASRU2011} and converted to a confusion network for compact representation using SRILM\cite{Stolcke02srilm--}. We used the pre-trained ASpIRE Chain Model which has been trained on Fisher English to transcribe the spoken question and extract the ASR lattices. The training dataset has a WER of $22.94\%$ and test set has a WER of $37.57\%$ on the best hypothesis of the ASR, while the cross-dataset evaluation test sets- NewsQA has a WER of $34.60\%$ and Freebase has a WER of $43.80\%$. 

\section{Experiments and Results}
We built our system over OpenNMT-Py\cite{opennmt}. We used a batch size of 32, dropout rate of 0.5, RNN size of 512 and decay steps 10000. The maximum number of parallel arcs in the confusion network and maximum sentence length are set to 20 and 50 respectively. The confusion network contains noise and interjections such as \emph{*DELETE*} and \emph{[noise]}, \emph{[laughter]}, \emph{uh}, \emph{oh} which leads to degradation in system performance. To mitigate the effect of such noise, we remove the whole set of parallel arcs if all the arcs are noise and interjection words. As shown in table \ref{tab:scores}, the pruned confusion network, named clean confnet, outperforms the system marginally for the SQuAD/HarvestingQA dataset. We also compare the system with a model trained on the best hypothesis of the extracted from the ASR lattice using Kaldi. Here, the confusion network encoder is replaced with a text encoder which shares weights with the factoid answer encoder. 

As shown in table \ref{tab:scores}, we observe for SQuAD/HarvestingQA dataset that the Best-ASR-hypothesis outperforms the clean confusion network model with a 5\% margin in BLEU score and 2\% margin in ROGUE-L score. To asses the cross-domain generlizability, we also perform cross-dataset evaluation by evaluating our models on $840$ samples of a KB based dataset(Freebase) and $470$ samples of a machine comprehension dataset(NewsQA). The clean confusion network marginally outperforms the best-hypothesis model in ROGUE scores for cross-dataset evaluation and gives comparable results on BLEU scores. This shows that the confusion network system generalizes better on cross-domain noisy data and is less sensitive to noise introduced by new domains and noisy input signal, when compared with the Best-ASR-Hypothesis model. A plausible reason to this could be that the confusion network model is itself trained on a closed set of hypothesis, as compared to the Best-ASR-Hypothesis model which makes simplifying assumptions about the input signal. A compelling extension to the confusion network model is to adapt the copy attention over all the time-aligned hypotheses of the confusion network input. This would allow the confusion network model to copy among top-N words at any given time-step of the confusion network, instead of an erroneous word with the highest ASR score.

An example of results on a SQuAD/HarvestingQA test sample is as follows:
\begin{table}[t!]
\centering
%     \begin{adjustbox}{width=\textwidth,center}
    % \begin{adjustbox}{center}
        \begin{tabular}{|c|l|l|l|l|l|}
        \hline
\textbf{Test Dataset} & \textbf{Input} & \textbf{BLEU} & \textbf{ROGUE-1} & \textbf{ROGUE-2} & \textbf{ROGUE-L} \\
        \hline
        \multirow{4}{*}{SQuAD/HarvestingQA} &
        \multicolumn{1}{|c|}{Best Hypothesis} &
        \multicolumn{1}{|c|}{\textbf{60.26}} & 
        \multicolumn{1}{|c|}{\textbf{82.43}} &
        \multicolumn{1}{|c|}{\textbf{70.61}} &
            \multicolumn{1}{|c|}{\textbf{78.21}} 
         \\\cline{2-6} &
         \multicolumn{1}{|c|}{Confnet} & 
         \multicolumn{1}{|c|}{55.38} &
         \multicolumn{1}{|c|}{81.60} &
         \multicolumn{1}{|c|}{68.02} &
         \multicolumn{1}{|c|}{76.68}                  \\\cline{2-6} &
         \multicolumn{1}{|c|}{Clean Confnet} &        \multicolumn{1}{|c|}{55.92} &
        \multicolumn{1}{|c|}{81.39} &
        \multicolumn{1}{|c|}{67.79} &
        \multicolumn{1}{|c|}{76.78}   
            \\\hline \hline
            \multirow{4}{*}{Freebase} &
            \multicolumn{1}{|c|}{Best Hypothesis} &
            \multicolumn{1}{|c|}{\textbf{43.21}}  & \multicolumn{1}{|c|}{71.37} &
            \multicolumn{1}{|c|}{51.72} &
            \multicolumn{1}{|c|}{64.98} 
            \\\cline{2-6} &
            \multicolumn{1}{|c|}{Confnet} &
            \multicolumn{1}{|c|}{41.86} &
            \multicolumn{1}{|c|}{72.42} &
            \multicolumn{1}{|c|}{51.84} &
            \multicolumn{1}{|c|}{\textbf{65.78}} 
            \\\cline{2-6} &
            \multicolumn{1}{|c|}{Clean Confnet} &
            \multicolumn{1}{|c|}{42.89} &
            \multicolumn{1}{|c|}{\textbf{72.54}} &
            \multicolumn{1}{|c|}{\textbf{52.77}} &
            \multicolumn{1}{|c|}{66.39}
            \\\hline \hline
            \multirow{4}{*}{NewsQA} & \multicolumn{1}{|c|}{Best Hypothesis} & 
            \multicolumn{1}{|c|}{49.98} &
            \multicolumn{1}{|c|}{75.82} &
            \multicolumn{1}{|c|}{59.59} &
            \multicolumn{1}{|c|}{72.65} 
            \\\cline{2-6} &
            \multicolumn{1}{|c|}{Confnet} &
            \multicolumn{1}{|c|}{53.45} &
            \multicolumn{1}{|c|}{\textbf{76.45}} &
            \multicolumn{1}{|c|}{60.32} &
            \multicolumn{1}{|c|}{72.78} 
            \\\cline{2-6} &
            \multicolumn{1}{|c|}{Clean Confnet} &
            \multicolumn{1}{|c|}{\textbf{56.86}} &
            \multicolumn{1}{|c|}{76.07} &
            \multicolumn{1}{|c|}{\textbf{61.18}} &
            \multicolumn{1}{|c|}{\textbf{73.12}} 
            \\\hline
        \end{tabular}
%     \end{adjustbox}
%     \vspace{ - 05 mm}
\caption{Top section shows the scores on 1000 SQuAD/HarvestingQA test samples. Bottom 2 section shows the scores for cross-dataset evaluation on a Knowledge-Base(Freebase) dataset and a machine comprehension(NewsQA) dataset. For each section, the top row displays the score on the best hypothesis of the confusion network, the middle row displays the scores on the confusion network, while the bottom row displays the results on the pruned clean confusion network}
    \label{tab:scores}
\end{table}
\begin{itemize}
    \item \textbf{Gold Question:}  what was the title of the sequel to conan the barbarian ?
    \item \textbf{Top-Hypothesis:} what was the title of the sequels are counting the barbarian
    \item \textbf{Factoid Answer:} conan the destroyer
    \item \textbf{Full-length Answer:} the title of the sequel to conan the barbarian was conan the destroyer
    \item \textbf{Clean Confnet Model prediction:} the title of the sequels to the barbarian was conan the destroyer
    \item \textbf{Best-Hypothesis Model prediction:} the title of the sequels are counting the barbarian
\end{itemize}

\section{Conclusion}
We propose the task of generating full-length natural answers from spoken questions and factoid answer. We generated a dataset consisting of triples (spoken question, factoid answer, full length answer) and extracted confusion network from the questions. We have used the pointer-network over ASR graphs(confusion network) and show that it gives comparable results to the model trained on the best hypothesis. Our system achieves a BLEU score of 55.92\% and ROGUE-L score of 76.78\% on SQuAD/HarvestingQA dataset. We perform cross-dataset evaluation to obtain a BLEU score of 42.89\% and ROGUE-L score of 66.39\% on Freebase, and a BLEU score of 56.86\% and ROGUE-L score of 73.12\% on NewsQA dataset.

%\section{Future Work}
%  We have used ASR scores of the confusion network as the probability distribution over which to copy the time-aligned hypothesis of the confusion network. However, to mitigate the ASR errors further, the copy-distribution of the confusion network can be learnt as a network parameter. We leave this for future work.

% %
% % ---- Bibliography ----
% %
% % BibTeX users should specify bibliography style 'splncs04'.
% % References will then be sorted and formatted in the correct style.
% %
\bibliographystyle{splncs04}
\bibliography{tsd2020}

\begin{thebibliography}{10}
\providecommand{\url}[1]{\texttt{#1}}
\providecommand{\urlprefix}{URL }
\providecommand{\doi}[1]{https://doi.org/#1}

\bibitem{DBLP:journals/csl/Hakkani-TurBRT06}
Hakkani{-}T{\"{u}}r, D., B{\'{e}}chet, F., Riccardi, G., T{\"{u}}r, G.: Beyond
  {ASR} 1-best: Using word confusion networks in spoken language understanding.
  Comput. Speech Lang.  \textbf{20}(4),  495--514 (2006).
  \doi{10.1016/j.csl.2005.07.005},
  \url{https://doi.org/10.1016/j.csl.2005.07.005}

\bibitem{DBLP:journals/corr/JagfeldV17}
Jagfeld, G., Vu, N.T.: Encodingword confusion networks with recurrent neural
  networks for dialog state tracking. CoRR  \textbf{abs/1707.05853} (2017),
  \url{http://arxiv.org/abs/1707.05853}

\bibitem{opennmt}
Klein, G., Kim, Y., Deng, Y., Senellart, J., Rush, A.M.: Open{NMT}: Open-source
  toolkit for neural machine translation. In: Proc. ACL (2017).
  \doi{10.18653/v1/P17-4012}, \url{https://doi.org/10.18653/v1/P17-4012}

\bibitem{Ladhak2016LatticeRnnRN}
Ladhak, F., Gandhe, A., Dreyer, M., Mathias, L., Rastrow, A., Hoffmeister, B.:
  Latticernn: Recurrent neural networks over lattices. In: INTERSPEECH (2016)

\bibitem{DBLP:journals/corr/abs-1808-02280}
Lee, C., Wang, S., Chang, H., Lee, H.: {ODSQA:} open-domain spoken question
  answering dataset. CoRR  \textbf{abs/1808.02280} (2018),
  \url{http://arxiv.org/abs/1808.02280}

\bibitem{DBLP:journals/corr/abs-1804-00320}
Li, C., Wu, S., Liu, C., Lee, H.: Spoken squad: {A} study of mitigating the
  impact of speech recognition errors on listening comprehension. CoRR
  \textbf{abs/1804.00320} (2018), \url{http://arxiv.org/abs/1804.00320}

\bibitem{Liu-et-al:0018ijcai}
Liu, C., He, S., Liu, K., Zhao, J.: Curriculum learning for natural answer
  generation. In: {IJCAI}. pp. 4223--4229. ijcai.org (2018)

\bibitem{8462030}
{Masumura}, R., {Ijima}, Y., {Asami}, T., {Masataki}, H., {Higashinaka}, R.:
  Neural confnet classification: Fully neural network based spoken utterance
  classification using word confusion networks. In: 2018 IEEE International
  Conference on Acoustics, Speech and Signal Processing (ICASSP). pp.
  6039--6043 (2018)

\bibitem{masamura-et-al-2018}
Masumura, R., Ijima, Y., Asami, T., Masataki, H., Higashinaka, R.: Neural
  confnet classification: Fully neural network based spoken utterance
  classification using word confusion networks. In: 2018 {IEEE} International
  Conference on Acoustics, Speech and Signal Processing, {ICASSP} 2018. pp.
  6039--6043 (04 2018). \doi{10.1109/ICASSP.2018.8462030}

\bibitem{pal2020modeling}
Pal, V., Guillot, F., Renders, J.M., Besacier, L.: Modeling asr ambiguity for
  dialogue state tracking using word confusion networks (2020)

\bibitem{pal-etal-2019-answering}
Pal, V., Shrivastava, M., Bhat, I.: Answering naturally: Factoid to full length
  answer generation. In: Proceedings of the 2nd Workshop on New Frontiers in
  Summarization. pp.~1--9. Association for Computational Linguistics, Hong
  Kong, China (Nov 2019). \doi{10.18653/v1/D19-5401},
  \url{https://www.aclweb.org/anthology/D19-5401}

\bibitem{Pennington14glove:global}
Pennington, J., Socher, R., Manning, C.D.: Glove: Global vectors for word
  representation. In: In EMNLP (2014)

\bibitem{Povey_ASRU2011}
Povey, D., Ghoshal, A., Boulianne, G., Burget, L., Glembek, O., Goel, N.,
  Hannemann, M., Motlicek, P., Qian, Y., Schwarz, P., Silovsky, J., Stemmer,
  G., Vesely, K.: The kaldi speech recognition toolkit. In: IEEE 2011 Workshop
  on Automatic Speech Recognition and Understanding. IEEE Signal Processing
  Society (Dec 2011), iEEE Catalog No.: CFP11SRW-USB

\bibitem{DBLP:journals/corr/SeeLM17}
See, A., Liu, P.J., Manning, C.D.: Get to the point: Summarization with
  pointer-generator networks. CoRR  \textbf{abs/1704.04368} (2017),
  \url{http://arxiv.org/abs/1704.04368}

\bibitem{Stolcke02srilm--}
Stolcke, A.: Srilm--an extensible language modeling toolkit. In: Proceedings of
  the 7th International Conference on Spoken Language Processing (ICSLP 2002.
  pp. 901--904 (2002)

\bibitem{unlu-inproceedings}
Unlu, M., Arisoy, E., Saraclar, M.: Question answering for spoken lecture
  processing. pp. 7365--7369 (05 2019). \doi{10.1109/ICASSP.2019.8682580}

\bibitem{zhong2018global}
Zhong, V., Xiong, C., Socher, R.: Global-locally self-attentive encoder for
  dialogue state tracking. In: ACL (2018)

\end{thebibliography}
% %\begin{thebibliography}{8}
% %\bibitem{ref_article1}
% %{Author, F.: Article title. Journal \textbf{2}(5), 99--110 (2016)}

% %\bibitem{ref_url1}
% %{LNCS Homepage, \url{http://www.springer.com/lncs}. Last accessed 4 Oct 2017}
% %\end{thebibliography}

\end{document}